%% file: sample-sigconf-authordraft.tex
\documentclass[manuscript]{acmart}
\AtBeginDocument{%
  }

\setcopyright{acmlicensed}
\copyrightyear{2018}
\acmConference[]{}{}{}
\acmBooktitle{}
\acmISBN{}

\usepackage[table]{xcolor}
\usepackage{multirow} 
\usepackage{mdframed} 
\newmdenv[
  linecolor=black,
  backgroundcolor=gray!20, 
  frametitlebackgroundcolor=gray!20, 
  frametitlerule=false,
  leftmargin=10pt,
  rightmargin=10pt,
  innerleftmargin=10pt,
  innerrightmargin=10pt,
  innertopmargin=10pt,
  innerbottommargin=10pt
]{myframe}

\begin{document}

\title{Mitigating the Threshold Priming Effect in Large Language Model–Based Relevance Judgments via Personality Infusing}

\author{Nuo Chen$\ast$}
\email{pleviumtan@outlook.com}
\affiliation{%
  \institution{The Hong Kong Polytechnic University}
  \state{HK}
  \country{China}
}

\author{Hanpei Fang$\ast$}
\email{hanpeifang@ruri.waseda.jp}
\affiliation{%
  \institution{Waseda University}
  \city{Tokyo}
  \country{Japan}
}

\author{Jiqun Liu}
\email{jiqunliu@ou.edu}
\affiliation{%
  \institution{The University of Oklahoma}
  \state{OK}
  \country{USA}
}

\author{Wilson Wei}
\email{w@eurexa.ai}
\affiliation{%
  \institution{EureXa Labs}
  \country{Singapore}
}


\author{Tetsuya Sakai}
 \email{tetsuyasakai@acm.org}
\affiliation{%
  \institution{Waseda University}
  \state{Tokyo}
  \country{Japan}
}

\author{Xiao-Ming Wu}
 \email{xiao-ming.wu@polyu.edu.hk}
\affiliation{%
  \institution{The Hong Kong Polytechnic University}
  \state{HK}
  \country{China}
}

\thanks{$\ast$ The first two authors contributed equally, the order is arranged in alphabetical order by surname. }

\renewcommand{\shortauthors}{Chen et al.}

\begin{abstract}
Recent research has explored LLMs as scalable tools for relevance labeling, but studies indicate they are susceptible to priming effects, where prior relevance judgments influence later ones. Although psychological theories link personality traits to such biases, it is unclear whether simulated personalities in LLMs exhibit similar effects. We investigate how Big Five personality profiles in LLMs influence priming in relevance labeling, using multiple LLMs on TREC 2021 and 2022 Deep Learning Track datasets. Our results show that certain profiles, such as High Openness and Low Neuroticism, consistently reduce priming susceptibility. Additionally, the most effective personality in mitigating priming may vary across models and task types. Based on these findings, we propose personality prompting as a method to mitigate threshold priming, connecting psychological evidence with LLM-based evaluation practices.
\end{abstract}





\received{20 February 2007}
\received[revised]{12 March 2009}
\received[accepted]{5 June 2009}

\maketitle

\input{sections/1_introduction}

\input{sections/2_rw}

\input{sections/3_approach}
\input{sections/4_experimental_setup}
\input{sections/5_result}
\input{sections/6_conclusion}

\section{Ethical Considerations}
This study does not involve human subjects or personally identifiable data. All experiments are conducted on publicly available datasets. We encourage future work to investigate safeguards and transparency mechanisms to ensure that personality conditioning is used responsibly and to minimize the risk of unintended harms.



\bibliographystyle{ACM-Reference-Format}
\bibliography{sample-base}



\end{document}

%% file: sections/1_introduction.tex
\section{Introduction}

Large language models (LLMs) are increasingly employed to automate evaluation in information retrieval (IR) systems, particularly for generating relevance labels. Yet, trained on human data, LLMs may inherit and amplify human biases, potentially yielding skewed models and suboptimal algorithms. Biased outputs can further contaminate future training corpora, creating a self-reinforcing cycle of bias~\citep{omiye2023large}. While most work has focused on social biases\citep{wang2023improving}, cognitive biases remain comparatively underexplored\citep{omiye2023large,Zhao2024gender,itzhak2023instructed,eicher2024reducing,schmidgall2024addressing,echterhoff-etal-2024-cognitive,lyu2025cognitivedebiasinglargelanguage}.

\textit{Cognitive biases} are systematic deviations in reasoning under uncertainty~\citep{kruglanski1983}, often leading to irrational judgments and suboptimal decisions~\citep{Tversky1974,Tversky1992}. In IR, numerous studies have investigated biases such as anchoring and priming and their impact on search behavior, satisfaction, and relevance judgments~\citep[e.g.][]{azzopardi21,liu2020reference,chen2025decoy,tyler2022,kelly2008,shokouhi-2015,eickhoff2018,scholer2013,liu2025boundedly,wang2024understanding,chen2022,chen2023}. However, few studies have examined cognitive biases in LLM outputs~\citep{chen2024ap,fang2025large}.


In this study, we investigate how to mitigate the cognitive bias introduced by \textit{threshold priming} when LLMs perform batched relevance assessments. The priming effect is a psychological phenomenon in which exposure to certain \textit{stimuli} (such as words, images, or concepts) unconsciously influences subsequent behavior, judgments, and decisions~\citep{tulving1990priming,VANDERWART198467,herr1986consequences}.~\citet{scholer2013} demonstrated that when assessors evaluate document relevance continuously, the quality of previously assessed documents can act as a threshold, triggering the priming effect: if earlier documents are of relatively low quality, assessors tend to assign higher relevance scores to subsequent documents, and \textit{vice versa}. \citet{chen2024ap} observed that LLMs also exhibit similar biases when performing batched relevance assessments, but did not propose mitigation strategies. 

Psychological literature indicates that certain specific personality traits (e.g., high openness) can be generally less affected by cognitive biases including the priming effect~\citep{stanovich2008thinking,ingendahl2023,meier2006turning}, while recent studies in NLP have shown that LLMs can be guided through specific instructions to generate responses consistent with the corresponding personality traits~\citep[e.g.,][]{jiang2023,jiang2024personallm}. To bridge the above research gap, 
we explore the use of personality simulation to mitigate threshold priming in LLM relevance assessment.

We seek to address two \textbf{research questions (RQs)}: (1) \textbf{RQ1}: What is the overall performance of different simulated personalities on threshold priming mitigation across the evaluated models? (2) \textbf{RQ2}: How do different simulated personalities perform across models when evaluated on different task types?

We conduct experiments with multiple LLMs on TREC 2021 and 2022 Deep Learning Track datasets (hereafter referred to as TRDL21 and TRDL22)~\citep{craswell2022overview,craswell2023overview}. Our results indicate that certain profiles (e.g., high openness, low neuroticism) stably reduce priming susceptibility in relevance labeling. In addition, the personality that achieves the most consistent mitigation effect may vary between models. We further find that when performing relevance labeling under different task types~\cite{liu2021deconstructing}, the personality that most consistently mitigates the priming effect may also vary. High Openness performs better on exploration and exploitation queries, whereas Low Neuroticism is more suitable for known item tasks. 

Our main contribution is: to the best of our knowledge, this paper is the first in IR field to address on the feasibility of using personality simulation to mitigate cognitive biases in LLMs, with a particular focus on alleviating the threshold priming effect. 
In summary, this work not only provides a practical approach to improving the reliability of LLMs as evaluators but also opens a new research direction for applying psychological theories to bias mitigation in artificial intelligence.



%% file: sections/2_rw.tex
\section{Related Work}
\subsection{Personality and Cognitive Biases}
\textbf{Personality} encompasses the emotional dispositions, attitudes, and behaviors that shape individual decision-making~\citep{Ellouze2024}. Understanding these traits offers insights into judgment processes and informs strategies in decision-making~\cite{RUSTICHINI2016122,Jalajas_Pullaro_2018}. The  five-factor model (FFM), also known as \textit{the Big Five traits} model, provides a widely adopted taxonomy for describing personality traits~\citep{CostaMcCrae1999FFT, McCraeJohn1992FFM}. FFM conceptualizes personality along five dimensions:~(1)~\textit{Openness} to Experience, representing curiosity and receptivity to novel ideas and experiences; (2)~\textit{Conscientiousness}, denoting responsibility and attention to detail; (3)~\textit{Extraversion}, reflecting sociability and engagement with others; (4)~\textit{Agreeableness}, capturing trust, empathy, and cooperativeness; (5)~\textit{Neuroticism} (with low scores indicating emotional stability), capturing tendencies toward negative affect and emotional reactivity. \textbf{Cognitive biases}, which arise from limits in human cognitive capacity, especially when information cannot be fully gathered or processed~\citep{kruglanski1983}, can cause an individual's decisions in uncertain situations to systematically deviate from the expectations of rational decision-making models~\citep{Tversky1974, Tversky1991, Tversky1992}. 

\subsubsection{Links between The Big Five Personality Traits and Cognitive Biases}

Empirical work in psychology showed links between personality traits and cognitive biases~\citep{meier2006turning,melinder2020,ingendahl2023,stanovich2008thinking,schindler2021bayes}. For example, neuroticism is associated with stronger confirmation bias and loss aversion;
Openness generally reduces cognitive biases, including confirmation bias, the anchoring effect and the priming effect, through greater cognitive flexibility~\citep{stanovich2008thinking,ingendahl2023,meier2006turning}. 

\subsubsection{Cognitive Biases in LLM responses}
Although LLMs lack human physiological structures, they may still acquire cognitive biases through exposure to data and human feedback during training and fine-tuning.~\citep{itzhak2023instructed}. There has been work studying various types of ssssscognitive biases 
in LLM generated responses~\citep{echterhoff-etal-2024-cognitive,schmidgall2024addressing,eicher2024reducing,itzhak2023instructed}. Some studies have also explored how to mitigate cognitive biases in LLM-generated responses~\citep{eicher2024reducing,he2025investigatingimpactllmpersonality}. However, these studies are limited to specific scenarios, such as university admission decisions~\citep{echterhoff-etal-2024-cognitive} or medical question answering~\citep{schmidgall2024addressing}, and how to mitigate biases such as threshold priming that arise when LLMs perform relevance judgments is awaiting exploration.  

\subsubsection{LLMs Simulating Personality}
There is a complementary line of research that leverages LLMs to simulate human traits in recent work. Several studies demonstrate that personality can be actively induced through carefully crafted prompts, persona conditioning, or chain-of-thought scaffolding, with models generating trait-congruent responses and narratives to standardized psychological inventories across repeated trials~\citep{jiang2023,jiang2024personallm,sorokovikova2024simulate,molchanova2025exploringpotentiallargelanguage}. Recent work further embeds such conditioned personas within agentic and robotic frameworks that incorporate memory, affect, and goals, allowing trait expression to persist across multi-turn interactions and dynamic contexts~\citep{park2024generative,lo2025llm}.



\subsection{Relevance Assessment}
Relevance assessment is essential in IR, both for building dataset to train rankers~\citep{liu2009} and for building test collections to evaluate rankers~\citep{sanderson2010,Voorhees-2002,sakai2023gold}.

\subsubsection{Cognitive Biases in Relevance Assessment}

In the IR community, some past research argued that human assessors can be influenced by a variety of cognitive biases when performing relevance assessments~\cite{liu2023behavioral}, and these biases may impact the consistency of the collected labels and affect evaluation reliability~\citep{shokouhi-2015, scholer2013, eickhoff2018}. 


\subsubsection{LLMs as Relevance Assessors}
Collecting relevance labels from human assessors is costly and time-consuming~\citep{bailey2008}. Consequently, large language models (LLMs) have recently been explored as automatic and scalable relevance assessors, demonstrating a consistent improvement in the alignment between human assessments and model predictions~\citep[e.g., ][]{faggioli23, thomas24, upadhyay2024umbrela,arabzadeh25,pires2025}. 
However, some studies argued that LLM assessors cannot fully replace human assessors~\citep{clarke2025llm,soboroff2025dont}, and that there are biases in the assessments made by LLM assessors~\citep{balog2025rankers}.  Recent studies have revealed that LLMs, when judging relevance, can exhibit similar biases~\citep{chen2024ap,fang2025large}. 

Our focus is not on the alignment between LLM relevance assessment and human assessments, but rather on mitigating the threshold priming bias in LLM assessment caused by context.



%% file: sections/3_approach.tex
\input{figures/methodology}
\section{Proposed Approach}
\textbf{Step 1: Personality Simulation/Infusing}. Inspired by~\citet{jiang2023}, we adopt an iterative approach, using the LLM to generate personality simulation instructions. For the Big Five personality traits, namely Openness, Conscientiousness, Extraversion, Agreeableness, and Neuroticism, we divide each dimension into ``high'' and ``low,'' resulting in a total of ten personality types: High Agreeableness (HA), Low Agreeableness (LA), High Conscientiousness (HC), Low Conscientiousness (LC), High Extraversion (HE), Low Extraversion (LE), High Neuroticism (HN), Low Neuroticism (LN), High Openness (HO), and Low Openness (LO). For each personality type, we submit the following instruction to an LLM.

\begin{quote}
\noindent{\emph{Please provide keywords related to \textcolor{blue}{\{personality\_type\}}}}
\end{quote}

After obtaining the keywords \textcolor{blue}{\{personality\_keywords\}} corresponding to the personality type, we use the following instruction to prompt the LLM to generate persona simulation instructions.

\begin{quote}
\noindent{\emph{ \textcolor{blue}{\{personality\_keywords\}}. Based on the keywords above, how would a person with \textcolor{blue}{\{personality\_type\}} behave when making judgments and decisions? Generate a prompt that instructs an LLM to imitate a person with \textcolor{blue}{\{personality\_type\}}.}}
\end{quote}

The upper panel of Figure~\ref{fig:method} illustrates the method we used to construct the personality simulation instructions. 
Using the above method, we obtained ten distinct personality simulation instructions. Including the default instruction (the empty string), we therefore obtained a total of eleven simulated personalities. For each personality, we perform the following batch assessment procedure.

\textbf{Step 2: Batch Assessment Procedure}
To compare how different threshold contexts affect LLMs’ relevance judgments on the same set of documents, we first sample a batch of documents (the epilogue) within a trial. We then sample a batch of highly relevant documents and a batch of low-relevance documents (the prologues), concatenate each with the epilogue, and perform batch relevance assessment on each combined set.

Inspired by~\citet{chen2024ap}, our procedure for conducting batch relevance assessment is as follows. For each trial, we first randomly select $n$ documents with a relevance score of $r_\mathrm{epilogue}$, to form the epilogue $\mathbf{E} = \{d^e_{1}, d^e_{2}, \ldots, d^e_{n}\}.$:

Each $d^e_i$ is a document with a relevance score of $r_\mathrm{epilogue}$.

Next, we randomly select $m$ documents with a groundtruth relevance of $r_\mathrm{low}$ and another $m$ documents with a groundtruth relevance score of $r_\mathrm{high}$ to form the low threshold (LT) prologue $\mathbf{L}$ and the high threshold (HT) prologue $\mathbf{H}$, respectively, where $r_\mathrm{low} < r_\mathrm{epilogue} < r_\mathrm{high} $:

$$
\mathbf{L} = \{d^l_{1}, d^l_{2}, \ldots, d^l_{m}\}, 
\mathbf{H} = \{d^h_{1}, d^h_{2}, \ldots, d^h_{m}\}.
$$
Each $d^l_i$ and is $d^h_i$ is a document with a ground truth relevance score of $r_\mathrm{low}$ and $r_\mathrm{high}$ respectively.

We then concatenate $\mathbf{L}$ with $\mathbf{E}$ and $\mathbf{H}$ with $\mathbf{E}$ to obtain the low threshold batch $\mathbf{B_l}$ and the high threshold batch $\mathbf{B_h}$, respectively:
$$
\mathbf{B_l} = \mathrm{concat}\left(\mathbf{L}, \mathbf{E}\right), 
\mathbf{B_h} = \mathrm{concat}\left(\mathbf{H}, \mathbf{E}\right).
$$

For $\mathbf{B_l}$ and $\mathbf{B_h}$, have an LLM assess the relevance of each document in batch, and output the results in JSON format. We used the instruction of ~\citet{chen2024ap} to prompt LLMs to conduct batch relevance assessment. 

Figure~\ref{fig:method} illustrates our approach. In this example, first, we prompt a LLM to generate keywords for the High Openness trait, then use these keywords to produce personality simulation instructions which are combined with the relevance assessment prompt to simulate assessors with different personalities performing the task. In the second step, the Epilogue contains the same set of documents for both high and low threshold batches. Following prior work~\citep{chen2024ap,scholer2013}, we compute the mean relevance scores under high (HT) and low (LT) thresholds and use their absolute difference $\Delta$ to quantify the priming effect. Smaller differences relative to the default indicate more effective mitigation.

%% file: figures/methodology.tex
\begin{figure}[htbp]
    \centering
    \includegraphics[width=0.99\linewidth]{
    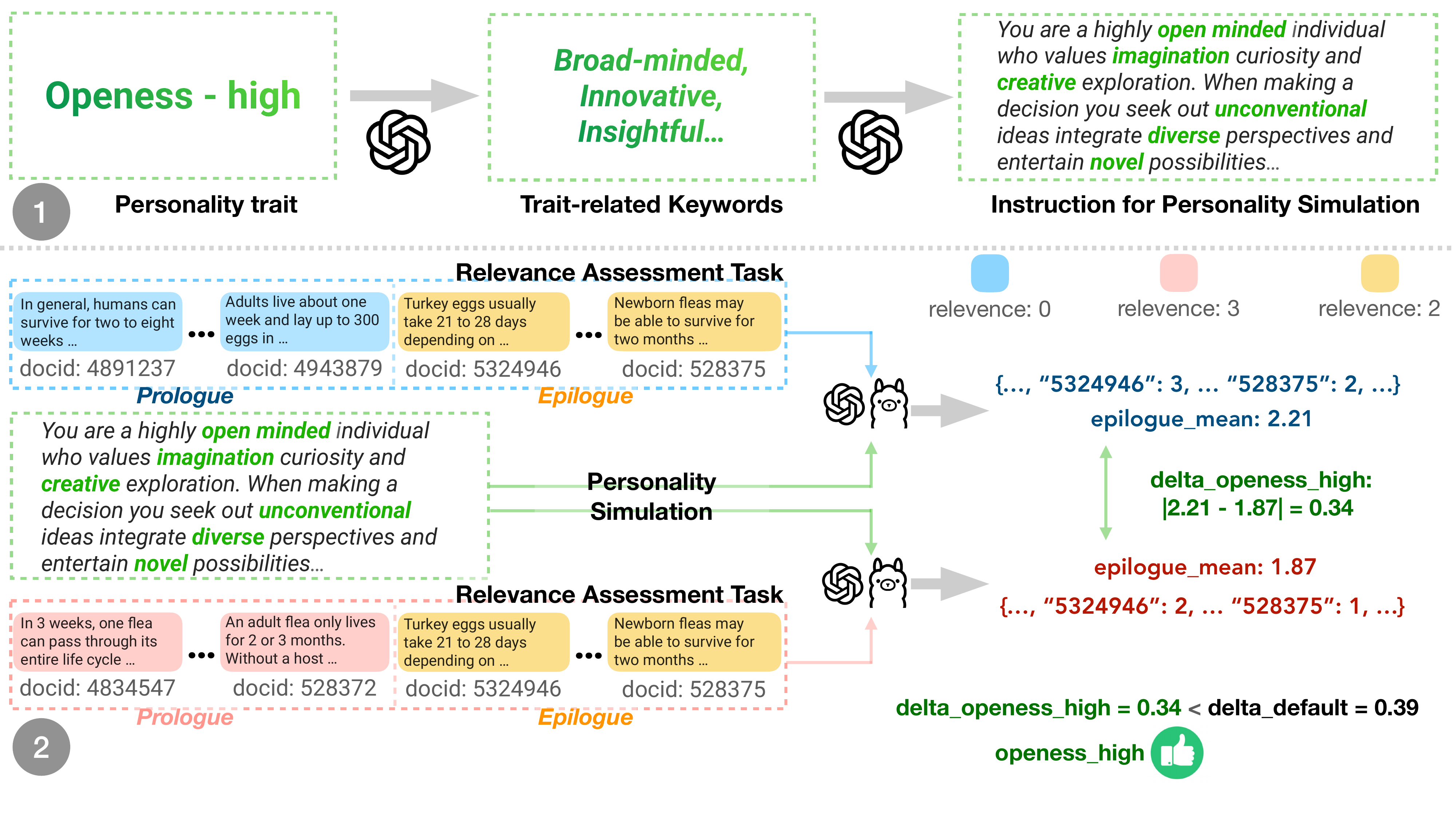
    }
    \caption{\label{fig:method} An example of the methodology adopted in our experiment.} 
    \Description{This is the illustration for our experiment}
\end{figure}

%% file: sections/4_experimental_setup.tex
\section{Experiments}
\subsection{Datasets and Evaluated LLMs}
 Our experiments are conducted on the passage retrieval test collections from the passage retrieval task of TRDL21 \cite{craswell2022overview} and TRDL22 \cite{craswell2023overview}. Following~\citet{chen2024ap}, we use only queries with relevance judgments, and exclude topics with fewer than 12 passages at any level of $\{0, 1, 2, 3\}$ to ensure adequate coverage across relevance levels. This results in a final selection of 15 TRDL21 topics and 34 TRDL22 topics, and 15+34 = 49 topics in total. For each topic, following the definitions of \citet{liu2021}, we used GPT o4 mini to categorize it as one of Known Item, Exploitation, or Exploration, and the assignments were manually modified. 17 topics were classified as Known Item, 16 topics were classified as Exploitation, and 16 topics were classified as Exploration.

We tested three LLMs, including two open-source models, LLaMA3-instruct-8B and LLaMA3-instruct-70B \cite{dubey2024llama}, and a proprietary model, GPT-3.5-turbo-1106. For comparability, identical decoding settings are applied to all LLMs: we set \texttt{top\_p} to $1.0$; and we set \texttt{temperature}, \texttt{frequency\_penalty} and \texttt{presence\_penalty} to $0$.

\subsection{Experimental Setup}

In our experiment, we use GPT-o4-mini to generate personality simulation instructions. For relevance assessment, we conduct 10 trials on each topic, and we set $r_\mathrm{low}=0$ and $r_\mathrm{high}=3$. To explore the extent to which LLMs are affected by the threshold priming effect when assessing documents of different relevance levels, we examined cases with $r_\mathrm{epilogue}=2$ and $r_\mathrm{epilogue}=1$. Additionally, to investigate how different batch lengths influence the magnitude of threshold priming in LLM assessments, we varied the batch length with different combinations of prologue and epilogue lengths, such as prologue length (PL) = 4 and epilogue length (EL) = 4, PL = 4 and EL = 8, PL = 8 and EL = 4, and PL = 8 and EL = 8. In summary, for each simulated personality (including the default), we conducted experiments on 39 topics, testing 4 batch lengths × 2 relevance levels = 8 distinct experimental configurations. For each configuration, we performed 10 independent trials and calculated the mean value of $\Delta$. If the mean $\Delta$ for a given personality is lower than that of the default, we consider that personality to mitigate threshold priming for that configuration. 

%% file: sections/5_result.tex
\section{Experimental Results}
Table~\ref{tab:personality-better-than-default} and \ref{tab:personality-performance-metrics} report, for each model and each personality, the number of cases (out of eight experimental configurations) where it outperforms the default. Bold indicates the best-performing personality for each model, and underline indicates the second-best. The light green background highlights the personality with the highest total number of improvements over the default.
\input{tables/overall_result}
\input{tables/by_topic}

\subsection{RQ1: Overall performance of different personalities across models}
Table~\ref{tab:personality-better-than-default} summarizes the number of configurations in which each simulated personality outperforms the default setting across three models. Each personality is evaluated on eight configurations, with higher counts indicating more consistent improvements. 

As shown in Table~\ref{tab:personality-better-than-default}, the mitigation effect of personality simulation varies across LLMs. For gpt-3.5-turbo, LN and HC are most effective, outperforming the default in 6/8 and 5/8 configurations, respectively. LN appears to reduce susceptibility to prior low-quality documents, while HC helps maintain adherence to relevance criteria, either mitigating threshold priming.

The Llama models show different optimal profiles. For llama-3-70b, HA, LC, and HO outperform the default in all eight configurations, indicating that cooperative, flexible, and cognitively adaptive traits are most effective. For llama-3-8b, HA and HN also achieve 8/8 improvements, underscoring that optimal profiles are model-specific.

Aggregated across models, LC and LN provide the most frequent improvements, with LC dominating the Llama models and LN leading for gpt-3.5-turbo. These results show that no single personality universally mitigates bias; profiles must be aligned with the target model. LC, LN, and HO consistently promote more stable and flexible decision-making, reducing over-adjustment to low-quality context and producing more consistent relevance judgments. The consistency across Llama models further suggests that these effects are independent of model size.


\subsection{RQ2: Performance of Different Personalities Across Models on Different Task Types}



Table \ref{tab:personality-performance-metrics} breaks down performance by task type—Known Item, Exploration, and Exploitation—showing that optimal personalities are both model- and task-dependent. G., L.70b, and L.8b denote gpt-3.5-turbo, llama-3-70B, and llama-3-8B, respectively.

\subsubsection{Known Item} For gpt-3.5-turbo, LN (6/8) and LO (5/8) perform best, suggesting that emotional stability and preference for established information yield the most reliable results for factual retrieval. In contrast, llama-3-70b and llama-3-8b benefit most from HA and HO, indicating that a cooperative and open-minded stance supports performance even in precise tasks.

\subsubsection{Exploration} Results shift notably, with HN and HO consistently leading across models. HN improves gpt-3.5-turbo in 7/8 cases and llama-3-8b in all 8, while HO excels especially on llama-3-70b (7/8). This suggests that HN’s cautious, detail-oriented nature and HO’s cognitive flexibility promote thorough exploration and reduce priming effects.

\subsubsection{Exploitation} For refining searches, gpt-3.5-turbo again benefits from LN (7/8) and HC (6/8), reinforcing that a stable, methodical persona best mitigates bias. llama-3-70b and llama-3-8b continue to favor HO and HN, respectively, showing consistent patterns across exploratory and exploitative settings.

Overall, Table \ref{tab:personality-performance-metrics}  highlights that no single personality is optimal. Effective bias mitigation requires aligning personality conditioning with both model and task: LN and HC for gpt-3.5-turbo on known-item and exploitation tasks, and HO/HN for Llama models and exploratory contexts. This points to a promising adaptive strategy where model personas are dynamically tuned to maximize reliable relevance assessments.

%% file: tables/overall_result.tex
\begin{table}[ht]
\caption{Number of experimental configurations where each simulated personality outperforms the default personality across three models. }
\centering
\begin{tabular}{cccc}
\toprule
\multirow{2}{*}{\textbf{Pers.}} & \multicolumn{3}{c}{\textbf{Number of Configurations Better than Default}} \\
 & \textbf{gpt-3.5-turbo} & \textbf{llama-3-70b} & \textbf{llama-3-8b} \\
\midrule
HA &  0/8  &  \textbf{8/8}  &  \textbf{8/8}  \\
LA  &  0/8  &  2/8  &  5/8  \\
HC &  \underline{5/8}  &  0/8  &  5/8   \\
\rowcolor{green!20} 
LC  &  3/8  &  \textbf{8/8}  &  \underline{7/8}   \\
HE &  3/8  &  5/8  &  4/8   \\
LE  &  2/8  &  2/8  &  4/8  \\
HN &  2/8  &  5/8  &  \textbf{8/8}   \\
\rowcolor{green!20} 
LN  &  \textbf{6/8}  &  \underline{6/8}  &  6/8   \\
HO &  2/8  &  \textbf{8/8}  &  \underline{7/8}   \\
LO  &  4/8  &  5/8  &  6/8   \\
\bottomrule
\end{tabular}
\label{tab:personality-better-than-default}
\end{table}

%% file: tables/by_topic.tex
\begin{table}[ht]
\caption{Number of experimental configurations where each simulated personality outperforms the default personality across three models, grouped by task types.}
\centering
\begin{tabular}{c ccc ccc ccc}
\toprule
\multirow{3}{*}{\textbf{P.}} 
& \multicolumn{9}{c}{\textbf{Number of Configurations Better than Default}} \\
\cmidrule(lr){2-10}
 & \multicolumn{3}{c}{\textbf{Known Item}} 
 & \multicolumn{3}{c}{\textbf{Exploration}} 
 & \multicolumn{3}{c}{\textbf{Exploitation}} \\
\cmidrule(lr){2-4}\cmidrule(lr){5-7}\cmidrule(lr){8-10}
 & \textbf{G.} & \textbf{L.70b} & \textbf{L.8b}
 & \textbf{G.} & \textbf{L.70b} & \textbf{L.8b}
 & \textbf{G.} & \textbf{L.70b} & \textbf{L.8b} \\
\midrule
HA & 0/8 & \textbf{8/8} & \textbf{8/8} & 1/8 & \underline{5/8} & 5/8 & 3/8 & \underline{6/8} & \underline{5/8} \\
LA & 0/8 & 4/8 & \underline{7/8} & 0/8 & 3/8 & 6/8 & 0/8 & 0/8 & 4/8 \\
HC & 2/8 & 3/8 & 6/8 & 5/8 & 2/8 & 5/8 & \underline{6/8} & 0/8 & 3/8 \\
LC & 2/8 & \textbf{8/8} & \underline{7/8} & 4/8 & \underline{5/8} & \underline{7/8} & 4/8 & 3/8 & 4/8 \\
HE & 3/8 & \underline{7/8} & 5/8 & 4/8 & 4/8 & 4/8 & 3/8 & 2/8 & 1/8 \\
LE & 2/8 & 5/8 & 5/8 & 1/8 & 4/8 & 4/8 & 2/8 & 0/8 & 2/8 \\
HN & 2/8 & 6/8 & \textbf{8/8} & \cellcolor{green!20}\textbf{7/8} & \cellcolor{green!20}\underline{5/8} & \cellcolor{green!20}\textbf{8/8} & 3/8 & 3/8 & \textbf{6/8} \\
LN & \cellcolor{green!20}\textbf{6/8} & \cellcolor{green!20}\underline{7/8} & \cellcolor{green!20}\underline{7/8} & 4/8 & 4/8 & 4/8 & \textbf{7/8} & 3/8 & 2/8 \\
HO & 0/8 & \textbf{8/8} & \underline{7/8} & \cellcolor{green!20}\underline{6/8} & \cellcolor{green!20}\textbf{7/8} & \cellcolor{green!20}\underline{7/8} & \cellcolor{green!20}2/8 & \cellcolor{green!20}\textbf{8/8} & \cellcolor{green!20}\textbf{6/8} \\
LO & \underline{5/8} & 6/8 & 6/8 & 4/8 & 4/8 & \underline{7/8} & 5/8 & 2/8 & 4/8 \\
\bottomrule
\end{tabular}

\label{tab:personality-performance-metrics}
\end{table}

%% file: sections/6_conclusion.tex
\section{Discussion and Conclusion}

This work presents the first systematic study of mitigating cognitive biases in LLM-based relevance assessment through personality conditioning. Using multiple LLMs across 49 TREC DL21 and DL22 topics, we show that conditioning on specific Big Five profiles significantly reduces susceptibility to threshold priming. Low Conscientiousness, Low Neuroticism, and High Openness yield the most consistent mitigation effects, though the optimal trait varies by model and task: High Openness is particularly effective in exploration and exploitation, whereas Low Neuroticism benefits known-item tasks. Nevertheless, a key limitation is that we do not address the prediction accuracy of LLM-based relevance judgments. Future work will explore combining personality conditioning with techniques such as continual pre-training to jointly improve accuracy and debiasing.